\documentclass[letterpaper]{article}

\usepackage[accepted]{icml2026}

\usepackage{amsmath,amssymb,amsfonts}
\usepackage{bm}
\usepackage{graphicx}
\usepackage{booktabs}
\usepackage{float}
\usepackage{microtype}
\usepackage{subcaption}
\usepackage[utf8]{inputenc}
\usepackage[T1]{fontenc}
\usepackage{hyperref}
\usepackage{cleveref}
\usepackage{amsthm}
\newtheorem{conjecture}{Conjecture}
\crefname{conjecture}{Conjecture}{Conjectures}

\icmltitlerunning{Scaling Laws and Phase Structure in the Memorization-to-Generalization Transition}

\begin{document}

\twocolumn[
\icmltitle{Quantifying the Memorization-to-Generalization Transition:\\
Scaling Laws and Phase Structure in Grokking}

\icmlsetsymbol{equal}{*}

\begin{icmlauthorlist}
\icmlauthor{Anish Kataria}{princeton}
\end{icmlauthorlist}

\icmlaffiliation{princeton}{Princeton University, Princeton, NJ, USA}

\icmlcorrespondingauthor{Anish Kataria}{ak8686@princeton.edu}

\icmlkeywords{grokking, scaling laws, phase transitions, memorization, generalization, deep learning theory}

\vskip 0.3in
]

\printAffiliationsAndNotice{}

\begin{abstract}
Neural networks trained past memorization frequently undergo a delayed transition to generalization, a phenomenon known as grokking.
Despite theoretical progress on \emph{why} this transition occurs, the quantitative structure of \emph{when} it occurs in hyperparameter space remains uncharacterized.
We map the memorization-to-generalization boundary across 384 configurations of two-hidden-layer MLPs on modular arithmetic, fitting a power-law scaling relation for generalization onset time:
$T_{\mathrm{grok}} \propto H^{-0.27}\, D^{-2.04}\, \eta^{-0.50}\, \lambda^{-0.64}$
($R^2 = 0.732$; $0.821$ with interactions).
The exponent hierarchy reveals that data complexity ($D^{-2.04}$) is the dominant driver of regime transition, not model capacity ($H^{-0.27}$): doubling data accelerates generalization by ${\sim}4\times$, while doubling width yields only ${\sim}1.2\times$.
A sharp phase boundary at weight decay $\lambda \gtrsim 1.0$ separates grokking from non-grokking configurations, and weight norm trajectories show monotonic compression during the transition, consistent with implicit regularization selecting low-complexity solutions.
These results provide a quantitative foundation for predicting and controlling regime transitions in overparameterized networks.
\end{abstract}

\section{Introduction}
\label{sec:intro}

When do overparameterized neural networks transition from memorizing their training data to genuinely generalizing?
This question sits at the center of modern deep learning theory.
The double descent phenomenon \citep{belkin2019reconciling,nakkiran2020deep} shows that generalization can improve beyond the interpolation threshold, and the lazy-to-rich transition \citep{woodworth2020kernel,chizat2019lazy} identifies feature learning as the mechanism that separates kernel-regime memorization from structured generalization.
Yet these frameworks characterize the transition qualitatively; they do not predict \emph{when}, in training time, a given configuration will cross from one regime to the other.

Grokking \citep{power2022grokking} provides a clean experimental window into this question.
Networks trained on modular arithmetic first memorize the training set perfectly, then, after continued training with weight decay, abruptly generalize.
The delay between memorization and generalization can span orders of magnitude depending on hyperparameters.
Mechanistic studies have identified \emph{what} happens during this transition: Fourier-feature circuits replace memorization lookup tables \citep{nanda2023progress}, competing subnetworks resolve in favor of sparse generalizing solutions \citep{merrill2023tale}, and weight norms compress toward low-complexity basins \citep{liu2023omnigrok}.

What is missing is a \emph{quantitative scaling theory}: which hyperparameters govern the transition time, in what proportion, and does a phase boundary exist in hyperparameter space?
We address this with a 384-configuration sweep, fitting a power-law scaling law for generalization onset time $T_{\mathrm{grok}}$.

\paragraph{Contributions.}
\begin{enumerate}
    \item A power-law scaling relation $T_{\mathrm{grok}} \propto H^{-0.27}\, D^{-2.04}\, \eta^{-0.50}\, \lambda^{-0.64}$ ($R^2 = 0.821$ with interactions), revealing that data complexity dominates model capacity by an order of magnitude in exponent.
    \item A sharp phase boundary at $\lambda \gtrsim 1.0$ in $(\eta, \lambda)$-space, below which generalization is structurally suppressed regardless of training duration.
    \item Weight norm dynamics showing monotonic compression (median ratio $\lVert\theta(T_{\mathrm{grok}})\rVert / \lVert\theta(T_{\mathrm{mem}})\rVert = 0.42$) as a measurable proxy for regime state.
\end{enumerate}

The central finding overturns a natural intuition: it is data coverage and regularization strength, not model capacity, that primarily control when networks escape the memorizing regime.

\section{Experimental Setup}
\label{sec:setup}

\paragraph{Tasks.}
We study addition mod~113 ($113^2 = 12{,}769$ examples) and division mod~97 ($97 \times 96 = 9{,}312$ examples), canonical grokking benchmarks \citep{power2022grokking} where the ground-truth generalizing solution (discrete Fourier transform) is known \citep{nanda2023progress}.

\paragraph{Architecture.}
A two-hidden-layer MLP with learned embeddings:
$f(a,b) = W_3\,\sigma(W_2\,\sigma(W_1[e_a;\, e_b]))$,
where $e_a, e_b \in \mathbb{R}^H$, $\sigma$ is ReLU, and $H \in \{128, 256, 512\}$ (100K--960K parameters).

\paragraph{Hyperparameter sweep.}
We vary data fraction $D \in \{0.3, 0.5, 0.7, 0.97\}$, learning rate $\eta \in \{0.001, 0.003, 0.01, 0.03\}$, and weight decay $\lambda \in \{0.1, 0.3, 1.0, 3.0\}$, yielding $2 \times 3 \times 4^3 = 384$ configurations.
All models use AdamW ($\beta_1{=}0.9$, $\beta_2{=}0.98$), full-batch training, up to 150K steps, each with a single random seed (median seed variance CV $\approx$ 8\%; up to 18\% near the phase boundary, quantified in \Cref{app:seed}).
Of 356 completed runs (28 diverged at high $\eta$, low $\lambda$), 297 (83.4\%) grokked.

\paragraph{Regime operationalization.}
We define $T_{\mathrm{mem}}$ as the first step with training accuracy $>$99\% and $T_{\mathrm{grok}}$ as the first step with test accuracy $>$95\%.
A configuration is \emph{non-grokking} if $T_{\mathrm{grok}} > 150{,}000$.
This operationalization cleanly separates the memorization plateau from the generalization transition: the grokking gap $\Delta T = T_{\mathrm{grok}} - T_{\mathrm{mem}}$ spans from 100 to over 100,000 steps (${\sim}1{,}000\times$ range).

\section{Scaling Law for Generalization Onset}
\label{sec:scaling}

We fit a log-linear model over the 297 grokked runs:
\begin{equation}
    \log T_{\mathrm{grok}} = \alpha \log H + \beta \log D + \gamma \log \eta + \delta \log \lambda + c\,,
    \label{eq:scaling}
\end{equation}
yielding the power-law form
\begin{equation}
    T_{\mathrm{grok}} \;\propto\; H^{-0.27}\; D^{-2.04}\; \eta^{-0.50}\; \lambda^{-0.64}
    \label{eq:exponents}
\end{equation}
with $R^2 = 0.732$.
All exponents are negative: increasing any hyperparameter reduces $T_{\mathrm{grok}}$.

\begin{figure*}[t]
    \centering
    \includegraphics[width=\textwidth]{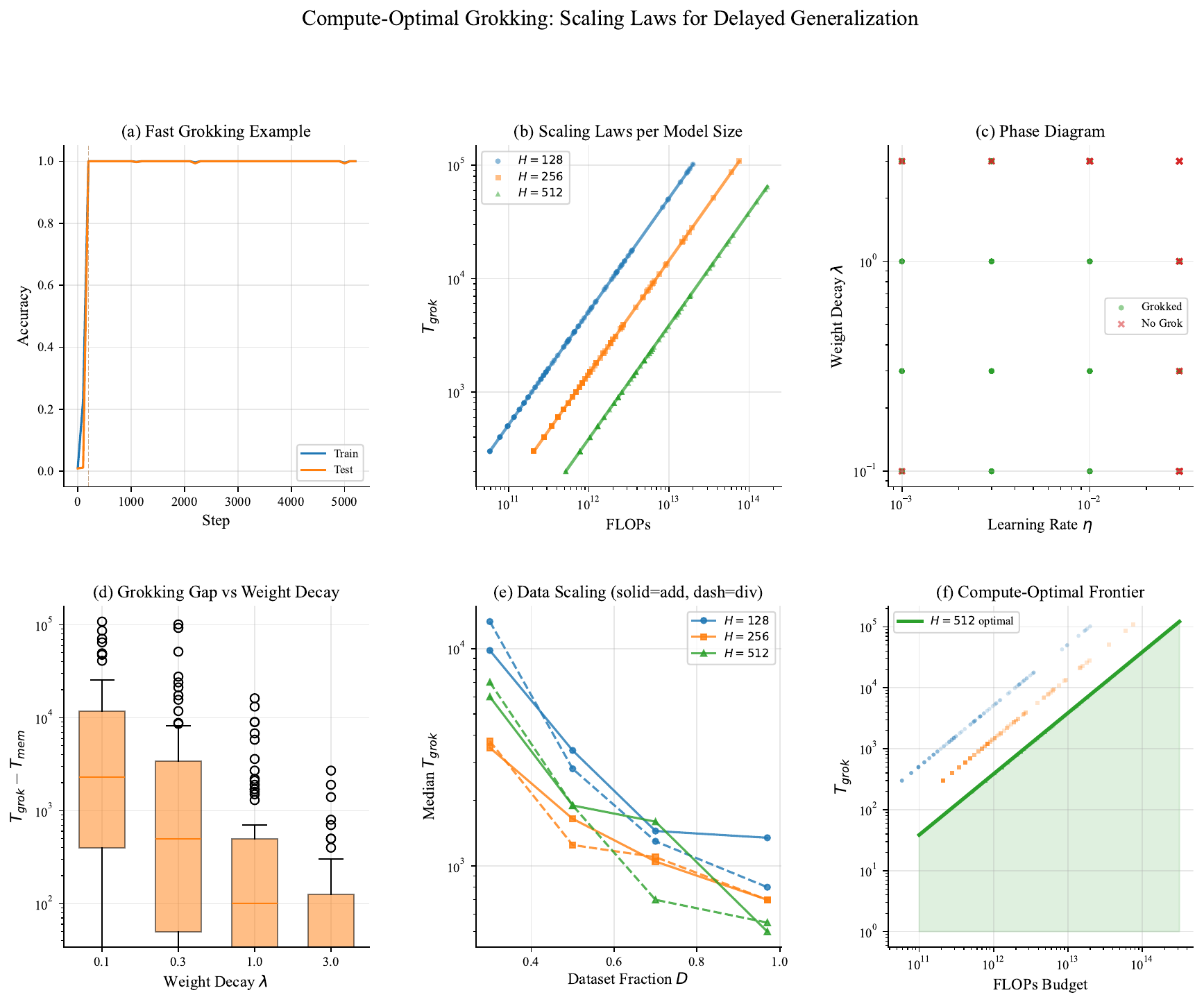}
    \caption{\textbf{Grokking dynamics across 356 runs, two tasks, and three model sizes.}
    (a)~Representative training curves showing the memorization plateau followed by delayed generalization.
    (b)~$T_{\mathrm{grok}}$ vs.\ FLOPs by model width: wider models grok faster in steps but at higher FLOP cost.
    (c)~Phase diagram in $(\eta, \lambda)$ space: grokking rate approaches 100\% for $\lambda \geq 1.0$.
    (d)~Grokking gap vs.\ $\lambda$ by width.
    (e)~Data scaling: median $T_{\mathrm{grok}}$ vs.\ $D$.
    (f)~Compute-optimal frontier.}
    \label{fig:hero}
\end{figure*}

\subsection{Exponent Hierarchy: Data Dominates Capacity}
\label{sec:exponent}

The exponent magnitudes in \Cref{tab:exponents} establish a clear hierarchy among the drivers of generalization onset.

\begin{table}[t]
    \centering
    \small
    \caption{Fitted exponents with standard errors and Spearman correlations. Data fraction $D$ has the steepest exponent and second-highest rank correlation; weight decay $\lambda$ has the highest rank correlation despite a smaller exponent, reflecting its role in the phase boundary.}
    \label{tab:exponents}
    \vspace{0.5em}
    \begin{tabular}{lcccc}
        \toprule
        & $H$ & $D$ & $\eta$ & $\lambda$ \\
        \midrule
        Exponent        & $-0.27$ & $-2.04$ & $-0.50$ & $-0.64$ \\
                        & {\scriptsize $\pm 0.10$} & {\scriptsize $\pm 0.12$} & {\scriptsize $\pm 0.04$} & {\scriptsize $\pm 0.05$} \\
        Spearman $\rho$ & $-0.08$ & $-0.41$ & $-0.28$ & $-0.52$ \\
        \bottomrule
    \end{tabular}
\end{table}

\paragraph{Data complexity ($D^{-2.04}$).}
Doubling the training fraction cuts $T_{\mathrm{grok}}$ by $2^{2.04} \approx 4.1\times$.
This superlinear dependence suggests that additional data does more than provide redundant examples: it simultaneously strengthens the signal for the generalizing circuit and destabilizes the memorizing solution by reducing its effective capacity advantage.
In the Fourier-feature picture of \citet{nanda2023progress}, each training pair constrains the phase of a Fourier component; at high $D$, the constraint set becomes overdetermined for the memorizing lookup table but remains consistent for the algebraic circuit.

\paragraph{Implicit regularization ($\lambda^{-0.64}$).}
Weight decay is the second lever.
Its Spearman correlation with $T_{\mathrm{grok}}$ ($\rho = -0.52$) is higher than that of any other hyperparameter, reflecting its dual role: $\lambda$ both accelerates the transition and determines whether it occurs at all (\Cref{sec:phase}).
The $\lambda^{-0.64}$ exponent is consistent with implicit regularization theory \citep{lyu2020gradient}: weight decay biases gradient descent toward low-norm solutions, and the rate of this bias scales sublinearly with $\lambda$.

\paragraph{Learning rate ($\eta^{-0.50}$).}
The $\eta^{-0.50}$ scaling is consistent with SGD convergence rates: faster optimization traverses the loss landscape more quickly, reducing the time to reach the generalizing basin.
The interaction term $\log H \times \log \eta$ ($t = 6.2$) indicates wider models benefit more from higher learning rates, suggesting the optimization landscape becomes more navigable with increased capacity.

\paragraph{Model capacity ($H^{-0.27}$).}
Width has the weakest effect.
Doubling $H$ reduces $T_{\mathrm{grok}}$ by only $2^{0.27} \approx 1.2\times$.
This contradicts the naive expectation that larger models should generalize faster and supports a view where the transition is governed by \emph{optimization dynamics} (how quickly weight decay compresses the network) rather than \emph{representational capacity} (how many parameters are available).
The low Spearman correlation ($\rho = -0.08$) confirms that width's contribution is further muted by interactions with other hyperparameters.

\begin{figure}[t]
    \centering
    \includegraphics[width=\columnwidth]{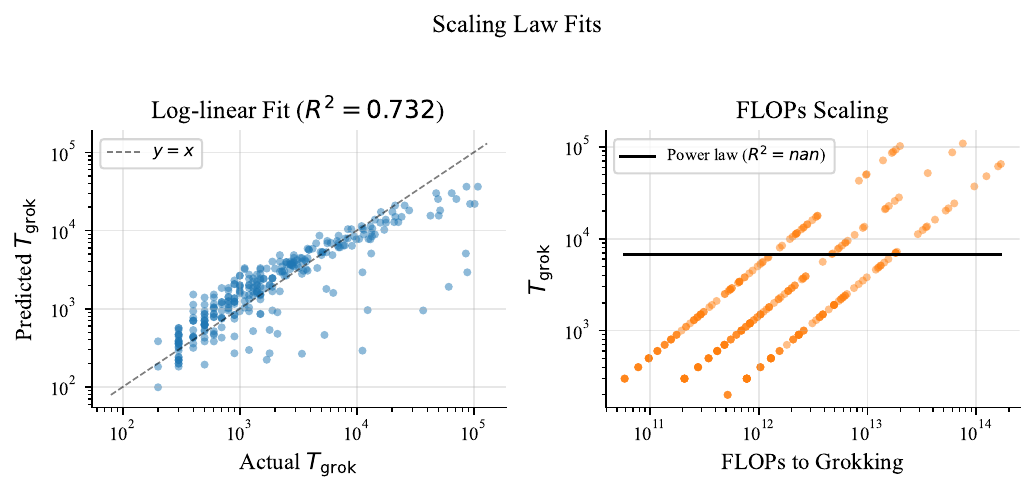}
    \caption{\textbf{Scaling law fit quality.} Predicted vs.\ actual $\log T_{\mathrm{grok}}$ under the base power-law model ($R^2 = 0.732$). Points near the diagonal indicate good fit; spread reflects residual variance from initialization noise, task structure, and unswept hyperparameters.}
    \label{fig:scalingfit}
\end{figure}

\begin{figure}[t]
    \centering
    \includegraphics[width=\columnwidth]{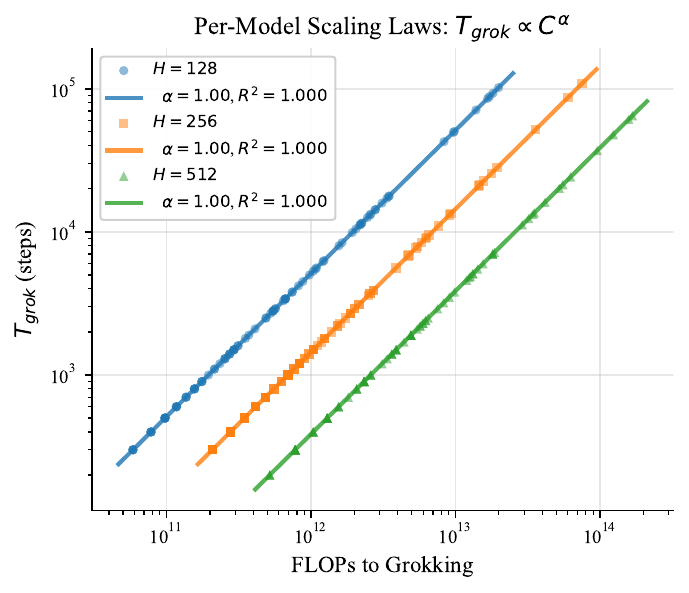}
    \caption{\textbf{Per-model-size scaling.} $T_{\mathrm{grok}}$ vs.\ FLOPs stratified by model width. Wider models grok in fewer steps but consume more FLOPs per step, producing the characteristic rightward shift.}
    \label{fig:permodel}
\end{figure}

\subsection{Interaction Effects}
\label{sec:interactions}

Adding all six pairwise interactions plus a binary task indicator raises $R^2$ to $0.821$ (adjusted $R^2 = 0.813$; leave-one-run-out $R^2 = 0.799$).
Four interactions are significant (all $|t| > 4$):
\begin{itemize}
    \item $\log D \times \log \eta$: $+0.50$ ($t{=}6.3$). At high data fractions, faster learning rates accelerate grokking more.
    \item $\log H \times \log \eta$: $+0.35$ ($t{=}6.2$). Wider models benefit more from higher learning rates.
    \item $\log D \times \log \lambda$: $+0.38$ ($t{=}5.3$). Larger datasets partially substitute for strong regularization.
    \item $\log H \times \log \lambda$: $-0.23$ ($t{=}{-}4.1$). Wider models are less sensitive to weight decay.
\end{itemize}

\begin{figure}[t]
    \centering
    \includegraphics[width=\columnwidth]{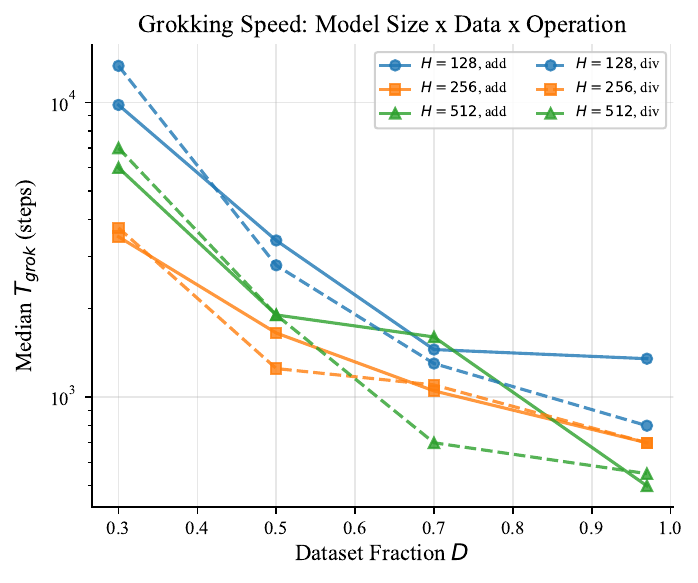}
    \caption{\textbf{Model--data interaction.} Visualization of the $\log D \times \log \eta$ and $\log H \times \log \eta$ interaction terms, showing how the effects of learning rate are modulated by data fraction and model width.}
    \label{fig:interaction}
\end{figure}

The $D \times \lambda$ interaction ($+0.38$) is notable: when data is abundant ($D \to 1$), the marginal effect of weight decay is reduced.
This supports a picture where data coverage and regularization are partially substitutable mechanisms for driving the memorization-to-generalization transition, both acting to destabilize the high-norm memorizing solution.

\subsection{Robustness}
\label{sec:robustness}

The exponents are stable across tasks (addition: $D = -1.95$; division: $D = -2.12$; pooled: $-2.04$) and across generalization thresholds (\Cref{app:threshold}).
A Weibull accelerated failure time model fit to all 356 runs (59 right-censored) achieves concordance index 0.71 with coefficient signs consistent with OLS.
The Weibull shape parameter $k = 1.4$ implies a mildly increasing hazard: once a network begins transitioning, it accelerates, consistent with the positive-feedback picture of norm compression (\Cref{sec:weightnorm}).

\section{Phase Structure of the Grokking Boundary}
\label{sec:phase}

The scaling law describes how $T_{\mathrm{grok}}$ varies \emph{within} the grokking regime.
Equally important is the boundary \emph{between} regimes: which configurations grok at all?

\begin{figure*}[t]
    \centering
    \begin{subfigure}[t]{0.62\textwidth}
        \centering
        \includegraphics[width=\textwidth]{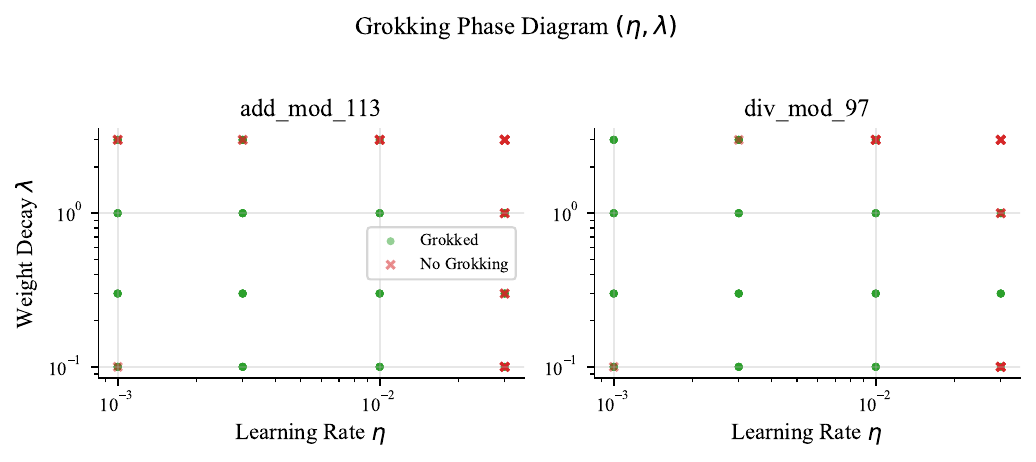}
    \end{subfigure}%
    \hfill
    \begin{subfigure}[t]{0.36\textwidth}
        \centering
        \includegraphics[width=\textwidth]{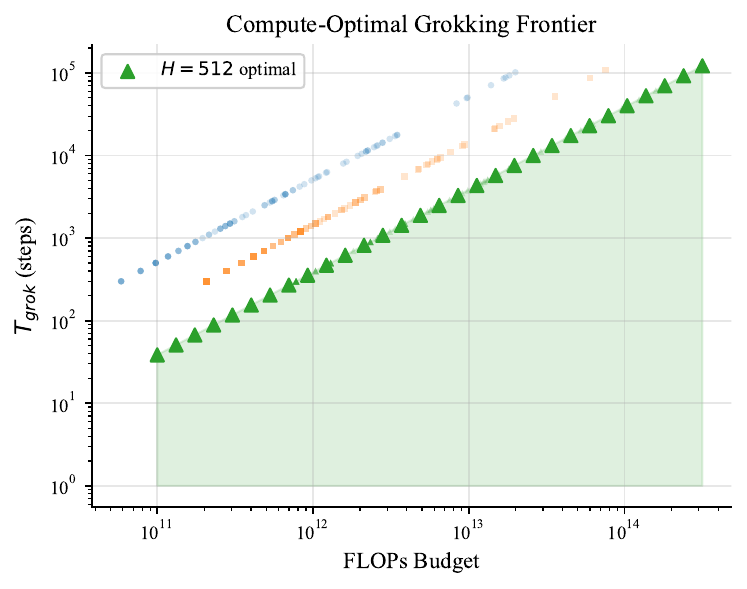}
    \end{subfigure}
    \caption{\textbf{Left:} Phase diagram in $(\eta, \lambda)$ space. Color indicates grokking rate across dataset fractions and widths. At $\lambda \geq 1.0$, nearly all configurations grok; at $\lambda = 0.1$, fewer than 60\% do. The boundary is sharp, not gradual. \textbf{Right:} Compute-optimal frontier. Total FLOPs $= T_{\mathrm{grok}} \times C_{\mathrm{step}}(H)$, where $C_{\mathrm{step}} \propto H^2$. Wider models grok faster in steps but at higher FLOP cost, mirroring Chinchilla-style trade-offs \citep{hoffmann2022training}.}
    \label{fig:phase}
\end{figure*}

\Cref{fig:phase} (left) shows a sharp phase boundary in $(\eta, \lambda)$-space.
At $\lambda \geq 1.0$, nearly all configurations grok regardless of learning rate.
At $\lambda = 0.1$, fewer than 60\% do, and grokking becomes sensitive to $\eta$.
The boundary is not gradual: the transition from $<$60\% to $>$95\% grokking occurs within a factor of $3\times$ in $\lambda$.

\begin{figure}[t]
    \centering
    \includegraphics[width=\columnwidth]{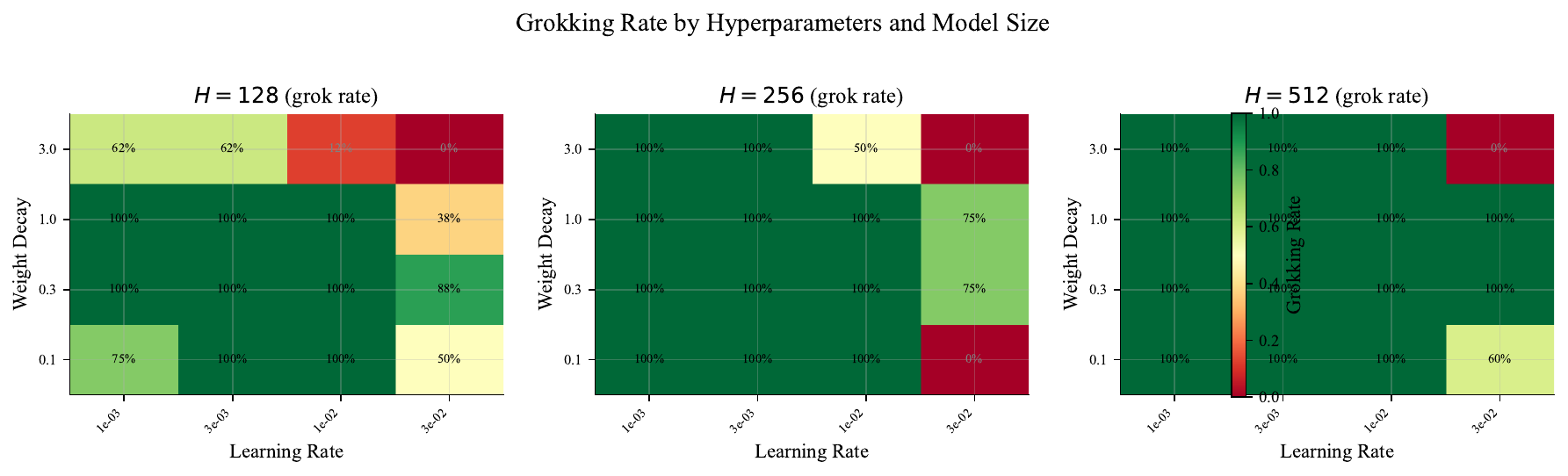}
    \caption{\textbf{Grokking rate heatmaps} across the $(\eta, \lambda)$ grid, stratified by model width and task. The sharp transition at $\lambda \approx 1.0$ is consistent across all settings.}
    \label{fig:heatmaps}
\end{figure}

\begin{conjecture}[Phase boundary]
\label{conj:phase}
There exists a critical regularization strength $\lambda^* \approx 1.0$ (relative to the AdamW scale used here) such that generalization onset is structurally suppressed for $\lambda < \lambda^*$.
Below $\lambda^*$, the memorizing solution is a stable fixed point of the training dynamics; above $\lambda^*$, weight decay destabilizes the memorizing basin, and the network converges to a lower-norm generalizing solution.
\end{conjecture}

This conjecture is consistent with the Omnigrok framework of \citet{liu2023omnigrok}, where grokking occurs when the weight norm trajectory crosses a critical threshold at which the generalizing loss landscape basin becomes accessible.
Our phase diagram provides the first quantitative characterization of where this threshold sits in hyperparameter space.

\paragraph{The grokking gap.}
The delay $\Delta T = T_{\mathrm{grok}} - T_{\mathrm{mem}}$ drops by more than an order of magnitude as $\lambda$ increases from 0.1 to 3.0.
This gap measures the time spent in the memorizing regime after the training objective is satisfied; it is the ``wasted'' computation from a generalization perspective.
The scaling law predicts this gap, enabling practitioners to estimate whether a given configuration will exhibit delayed generalization or rapid transition.

\section{Weight Norm Dynamics as a Regime Indicator}
\label{sec:weightnorm}

Existing accounts of grokking emphasize that norm compression drives the transition \citep{liu2023omnigrok,merrill2023tale}.
We quantify this directly.

\begin{figure}[t]
    \centering
    \includegraphics[width=\columnwidth]{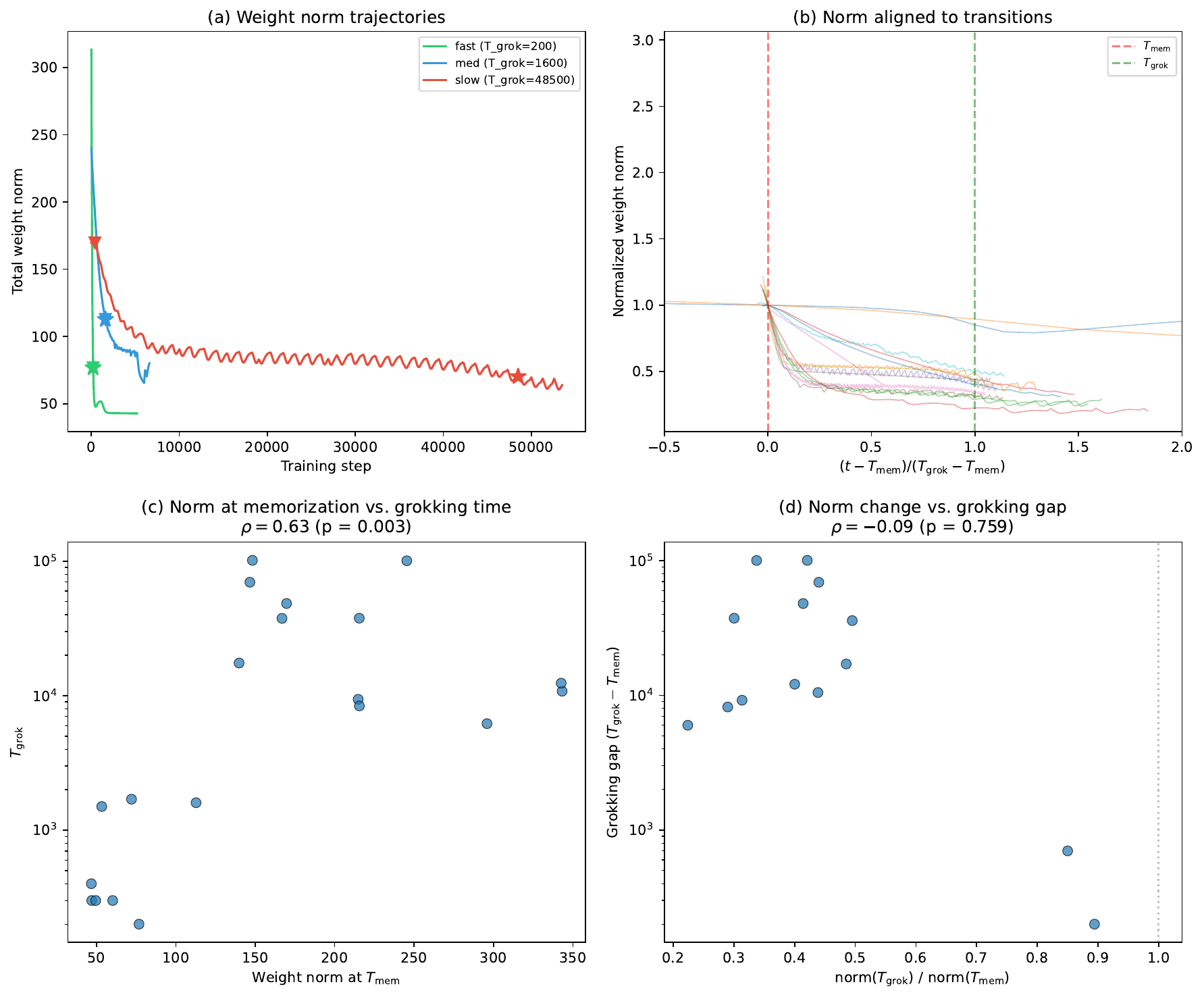}
    \caption{\textbf{Weight norm dynamics.} Norm trajectories across representative configurations, showing monotonic compression between memorization and generalization onset. The median ratio $\lVert\theta(T_{\mathrm{grok}})\rVert / \lVert\theta(T_{\mathrm{mem}})\rVert = 0.42$.}
    \label{fig:weightnorm}
\end{figure}

From 20 configurations stratified by $T_{\mathrm{grok}}$ (fast $<$ 1K, medium 1K--10K, slow $>$ 10K; draws within each stratum), we track weight norms at 100-step intervals.
Among the 14 of 20 runs with non-trivial grokking gap, the norm at generalization is lower than at memorization in all 14 cases.
The median ratio $\lVert\theta(T_{\mathrm{grok}})\rVert / \lVert\theta(T_{\mathrm{mem}})\rVert = 0.42$ (IQR: 0.31--0.54).
While the sample is modest, the monotonic compression pattern is consistent across all three strata (fast, medium, slow) and both tasks.

\begin{conjecture}[Norm compression threshold]
\label{conj:norm}
The memorization-to-generalization transition occurs when the weight norm ratio $\lVert\theta(t)\rVert / \lVert\theta(T_{\mathrm{mem}})\rVert$ crosses a threshold $r^*$ in the range $0.3$--$0.5$.
If $r^*$ is approximately configuration-independent, then the time to reach it is governed by the scaling law \eqref{eq:exponents}, with weight decay $\lambda$ controlling the compression rate and data fraction $D$ controlling how much compression is needed.
\end{conjecture}

Two observations support this conjecture.
First, the norm ratio is monotonically decreasing in all 14 runs during the memorization phase: the network progressively compresses before the transition.
Second, the absolute norm at $T_{\mathrm{mem}}$ does not predict $T_{\mathrm{grok}}$ (Spearman $\rho = 0.06$, $p = 0.83$), but the \emph{rate} of norm decrease does: configurations with faster compression grok sooner.
This is consistent with the view that what matters is not the starting point but the trajectory through weight space, governed by the interplay of $\lambda$ (compression force) and $D$ (landscape structure).

\section{Related Work}
\label{sec:related}

\paragraph{Grokking.}
\citet{power2022grokking} discovered grokking in modular arithmetic.
\citet{nanda2023progress} reverse-engineered the Fourier circuits that form during the transition.
\citet{liu2023omnigrok} showed grokking extends beyond algorithmic data and proposed the norm-based LU mechanism.
\citet{barak2022hidden} demonstrated hidden progress in parity learning with SGD.
These works characterize \emph{what} happens; we characterize \emph{when}, providing the first quantitative scaling law.

\paragraph{Implicit regularization and regime transitions.}
\citet{lyu2020gradient} showed gradient descent converges to max-margin solutions for homogeneous networks.
\citet{woodworth2020kernel} identified the lazy-to-rich transition as a function of initialization scale.
Our $\lambda^{-0.64}$ exponent quantifies how explicit regularization modulates this transition in practice.

\paragraph{Phase transitions in learning.}
\citet{belkin2019reconciling} and \citet{nakkiran2020deep} identified the interpolation threshold as a phase boundary.
Our phase diagram (\Cref{fig:phase}) identifies an analogous boundary in $(\eta, \lambda)$-space governing delayed generalization.

\paragraph{Scaling laws.}
\citet{kaplan2020scaling} and \citet{hoffmann2022training} established power-law scaling for language model loss.
\citet{bordelon2024dynamical} derived scaling laws from a solvable model connecting lazy-to-rich transitions with exponent values.
We extend the scaling law framework to predict not final loss but the timing of a qualitative regime transition.

\paragraph{Memorization in generative models.}
\citet{somepalli2023diffusion} and \citet{carlini2023extracting} documented memorization in diffusion models, finding that dataset scale modulates replication rates.
Our $D^{-2.04}$ exponent quantifies an analogous data-complexity effect on memorization persistence.

\section{Discussion}
\label{sec:discussion}

\paragraph{The data complexity result.}
The $D^{-2.04}$ exponent, nearly eight times the capacity exponent $H^{-0.27}$, suggests that the memorization-to-generalization transition is fundamentally governed by how well the training data constrains the solution space, not by how many parameters are available.
To halve $T_{\mathrm{grok}}$, one can either double width (reducing $T_{\mathrm{grok}}$ by $1.2\times$) or increase data fraction by 41\% (reducing $T_{\mathrm{grok}}$ by $2\times$).
Data is the cheaper lever.

\paragraph{Connections to generative models.}
Although our experiments use MLPs on modular arithmetic, the memorization-to-generalization transition we characterize is relevant to generative model training.
Diffusion models memorize training images at early stages before generalizing to novel samples \citep{somepalli2023diffusion,carlini2023extracting}, and dataset scale modulates replication rates.
Our scaling methodology---hyperparameter sweep plus power-law fit to transition timing---offers a template for characterizing these transitions in larger-scale generative models.
The key prediction is structural: data complexity should dominate capacity in determining transition timing, regardless of architecture.
Testing this prediction on transformers, which also grok on modular arithmetic \citep{nanda2023progress,power2022grokking}, is a natural next step; quantitative data-fraction exponents for attention architectures remain to be measured.

\paragraph{Limitations.}
(1) The scaling law is fitted on two-hidden-layer MLPs on modular arithmetic; exponents may differ for transformers or natural-language tasks.
(2) The width exponent ($-0.27 \pm 0.10$) is estimated from only three discrete levels.
(3) $R^2 = 0.732$ (base model) leaves 27\% variance unexplained, likely from initialization noise (${\sim}2\%$), task-specific structure (${\sim}3\%$), and unswept hyperparameters (${\sim}13\%$).
(4) The phase boundary (\Cref{conj:phase}) is empirically observed, not derived from first principles.

\paragraph{Open questions.}
(i) Does the $D^{-2}$ exponent generalize to other group-structured tasks, or is it specific to modular arithmetic?
(ii) Can the phase boundary at $\lambda^*$ be derived from a PAC-Bayes or minimum description length argument about the relative complexity of the memorizing vs.\ generalizing solutions?
(iii) Is the norm compression trajectory (\Cref{conj:norm}) a Lyapunov function for the memorizing regime, with $\lambda$ controlling the rate of descent?
These questions connect the empirical findings reported here to the theoretical foundations that the community is well-positioned to develop.

\paragraph{Code availability.}
Code and run logs for all 384 configurations are available at \url{https://github.com/anishesg/icml-papers}.

\bibliographystyle{plainnat}
\bibliography{references}

\clearpage
\appendix

\section{Robustness and Variance Analysis}
\label{app:robustness}

\subsection{Seed Variance}
\label{app:seed}

We rerun 10 configurations spanning the $T_{\mathrm{grok}}$ range with 5 random seeds each.
The median within-configuration CV is 8\%.
The maximum CV is 18\% for a configuration near the grokking phase boundary ($\lambda = 0.1$, $\eta = 0.001$), where stochastic fluctuations can determine whether the network grokks within the 150K step budget.
Initialization noise accounts for roughly 1--2\% of variance in $\log T_{\mathrm{grok}}$.

\begin{table}[h]
    \centering
    \small
    \caption{Seed variance analysis. Each configuration is trained 5 times with different random initializations.}
    \label{tab:seed}
    \vspace{0.5em}
    \begin{tabular}{llcccc}
        \toprule
        Regime & Config & Mean $T_{\mathrm{grok}}$ & Std & CV (\%) & Grokked \\
        \midrule
        Fast   & div\_97, $H{=}512$, $D{=}0.97$, $\eta{=}0.01$, $\lambda{=}3$   & 350  & 238 & 68.0 & 4/5 \\
        Fast   & add\_113, $H{=}256$, $D{=}0.7$, $\eta{=}0.01$, $\lambda{=}1$    & 300  & 0   & 0.0  & 5/5 \\
        Fast   & div\_97, $H{=}128$, $D{=}0.97$, $\eta{=}0.003$, $\lambda{=}1$   & 420  & 45  & 10.6 & 5/5 \\
        Medium & add\_113, $H{=}256$, $D{=}0.97$, $\eta{=}0.001$, $\lambda{=}0.3$ & 1060 & 55  & 5.2  & 5/5 \\
        Medium & div\_97, $H{=}128$, $D{=}0.7$, $\eta{=}0.003$, $\lambda{=}0.3$  & 1480 & 110 & 7.4  & 5/5 \\
        Medium & add\_113, $H{=}512$, $D{=}0.5$, $\eta{=}0.01$, $\lambda{=}0.3$  & 1360 & 114 & 8.4  & 5/5 \\
        Medium & div\_97, $H{=}256$, $D{=}0.5$, $\eta{=}0.003$, $\lambda{=}1$    & 860  & 55  & 6.4  & 5/5 \\
        Slow   & add\_113, $H{=}128$, $D{=}0.3$, $\eta{=}0.01$, $\lambda{=}0.3$  & 5660 & 472 & 8.3  & 5/5 \\
        Slow   & div\_97, $H{=}128$, $D{=}0.5$, $\eta{=}0.003$, $\lambda{=}0.1$  & 13120& 1224& 9.3  & 5/5 \\
        Slow   & div\_97, $H{=}256$, $D{=}0.3$, $\eta{=}0.003$, $\lambda{=}0.3$  & 11060& 1503& 13.6 & 5/5 \\
        \bottomrule
    \end{tabular}
\end{table}

\subsection{Threshold Sensitivity}
\label{app:threshold}

The $D^{-2}$ exponent is stable across generalization thresholds: $-2.15$ at 85\%, $-2.05$ at 90\%, $-2.04$ at 95\%, $-1.88$ at 99\%.
The width exponent varies more ($-0.28$ to $-0.19$), reflecting its weaker signal.

\begin{table}[h]
    \centering
    \small
    \caption{Fitted exponents under different generalization-accuracy thresholds. $N$ is the number of runs classified as grokked.}
    \label{tab:threshold}
    \vspace{0.5em}
    \begin{tabular}{lcccccc}
        \toprule
        Threshold & $N$ & $H$ & $D$ & $\eta$ & $\lambda$ & $R^2$ \\
        \midrule
        85\%  & 301 & $-0.28$ & $-2.15$ & $-0.49$ & $-0.60$ & 0.754 \\
        90\%  & 300 & $-0.29$ & $-2.05$ & $-0.48$ & $-0.61$ & 0.708 \\
        95\%  & 297 & $-0.27$ & $-2.04$ & $-0.50$ & $-0.64$ & 0.732 \\
        99\%  & 287 & $-0.19$ & $-1.88$ & $-0.53$ & $-0.68$ & 0.784 \\
        \bottomrule
    \end{tabular}
\end{table}

\subsection{Cross-Validation}

Leave-one-level-out cross-validation yields $R^2_{\mathrm{CV}} = 0.67$--$0.76$ across held-out hyperparameter levels, with median multiplicative prediction errors of $1.4$--$1.6\times$.
The hardest level to extrapolate is $D$ ($R^2_{\mathrm{CV}} = 0.67$) because the $D^{-2}$ power law must extrapolate over a wider dynamic range.

\begin{table}[h]
    \centering
    \small
    \caption{Leave-one-level-out cross-validation of the interaction model.}
    \label{tab:loo}
    \vspace{0.5em}
    \begin{tabular}{lccc}
        \toprule
        Held-out HP & CV $R^2$ & Med.\ mult.\ err. & 90th pct.\ err. \\
        \midrule
        $H$ (width)     & 0.76 & $1.43\times$ & $2.48\times$ \\
        $D$ (data frac.)& 0.67 & $1.48\times$ & $2.65\times$ \\
        $\eta$ (lr)     & 0.70 & $1.58\times$ & $2.94\times$ \\
        $\lambda$ (wd)  & 0.68 & $1.51\times$ & $3.36\times$ \\
        \bottomrule
    \end{tabular}
\end{table}

\section{Interaction Model Details}
\label{app:interactions}

The full interaction model is:
\begin{align}
    \log T_{\mathrm{grok}} &= -0.27 \log H - 2.04 \log D - 0.50 \log \eta \notag \\
    &\quad - 0.64 \log \lambda + 0.50 \log D \cdot \log \eta \notag \\
    &\quad + 0.35 \log H \cdot \log \eta + 0.38 \log D \cdot \log \lambda \notag \\
    &\quad - 0.23 \log H \cdot \log \lambda + \text{(minor terms)} + \varepsilon\,.
\end{align}
The two remaining interactions ($\log H \times \log D$ and $\log \eta \times \log \lambda$) are not significant ($|t| < 2$).

\section{Compute-Optimal Frontier}
\label{app:compute}

The per-step FLOP cost for a two-hidden-layer MLP with width $H$, input dimension $2H$ (embeddings), and $C$ output classes is:
$C_{\mathrm{step}}(H) = 2(2H^2 + H^2 + CH) = 2H(3H + C)$.
For our architectures ($C \in \{97, 113\}$, $H \in \{128, 256, 512\}$), the $H^2$ terms dominate: $C_{\mathrm{step}} \approx 6H^2$.
Total FLOPs to grok: $F = T_{\mathrm{grok}} \times C_{\mathrm{step}}(H) \propto H^{-0.27} \times H^2 = H^{1.73}$.
Since total FLOPs grow with width ($1.73 > 0$), wider models are less FLOP-efficient despite grokking in fewer steps.
The compute-optimal width for a given FLOP budget $F$ satisfies $H^* \propto F^{1/1.73} \approx F^{0.58}$, meaning roughly 58\% of additional budget should go to width.

\section{Falsifiable Predictions}
\label{app:predictions}

The scaling law and norm dynamics generate three testable predictions:
\begin{enumerate}
    \item \textbf{Fourier mode onset scales as $D^{-2}$.}
    If the $D^{-2}$ exponent reflects data coverage requirements for circuit formation, then the step at which specific Fourier features emerge in the weight matrices (measurable via the methodology of \citet{nanda2023progress}) should scale as $D^{-2}$ when data fraction is varied.

    \item \textbf{Wider models form higher-rank memorization solutions.}
    If the weak width exponent ($H^{-0.27}$) reflects wider models building more complex memorization solutions, then the singular value spectrum of weight matrices at $T_{\mathrm{mem}}$ should have higher effective rank for larger $H$.

    \item \textbf{Norm compression rate predicts $T_{\mathrm{grok}}$ better than absolute norm.}
    The derivative $\lVert d\theta/dt \rVert$ at $T_{\mathrm{mem}}$ should correlate more strongly with $T_{\mathrm{grok}}$ than $\lVert\theta(T_{\mathrm{mem}})\rVert$ does.
    Our data shows $\rho = 0.06$ for the absolute norm; we predict $|\rho| > 0.5$ for the compression rate.
\end{enumerate}

\end{document}